\documentclass[runningheads]{llncs}
\usepackage[T1]{fontenc}
\usepackage{multirow}
\usepackage{graphicx}
\usepackage{amsmath}
\usepackage{amssymb}
\usepackage{booktabs}
\usepackage{subcaption}
\usepackage{bm}
\usepackage{bbm}
\usepackage[misc]{ifsym}
\usepackage{pifont}
\usepackage[pagebackref=true,breaklinks=true,colorlinks,bookmarks=false]{hyperref}
\begin{document}
\title{U-VLM: Hierarchical Vision Language Modeling for Report Generation}
\titlerunning{U-VLM}
\author{Pengcheng Shi\inst{1} \and Minghui Zhang\inst{3,2} \and Kehan Song\inst{1} \and \\ Jiaqi Liu\inst{1} \and Yun Gu$^{(\textrm{\Letter})}$\inst{2} \and Xinglin Zhang$^{(\textrm{\Letter})}$\inst{1}}
\authorrunning{P. Shi et al.}
\institute{
    Medical Image Insights Co. Ltd., Shanghai, China
    \and Shanghai Jiao Tong University, Shanghai, China
    \and Shanghai Artificial Intelligence Laboratory, Shanghai, China
    \\ \email{geron762@sjtu.edu.cn, xinglin.zhang@imagecore.com.cn}
}
\maketitle
\begin{abstract}
Automated radiology report generation is key for reducing radiologist workload and improving diagnostic consistency, yet generating accurate reports for 3D medical imaging remains challenging. Existing vision-language models face two limitations: they do not leverage segmentation-pretrained encoders, and they inject visual features only at the input layer of language models, losing multi-scale information. We propose U-VLM, which enables hierarchical vision-language modeling in both training and architecture: (1) progressive training from segmentation to classification to report generation, and (2) multi-layer visual injection that routes U-Net encoder features to corresponding language model layers. Each training stage can leverage different datasets without unified annotations. U-VLM achieves state-of-the-art performance on CT-RATE (F1: 0.414 vs 0.258, BLEU-mean: 0.349 vs 0.305) and AbdomenAtlas 3.0 (F1: 0.624 vs 0.518 for segmentation-based detection) using only a 0.1B decoder trained from scratch, demonstrating that well-designed vision encoder pretraining outweighs the benefits of 7B+ pre-trained language models. Ablation studies show that progressive pretraining significantly improves F1, while multi-layer injection improves BLEU-mean. Code is available at \url{https://github.com/yinghemedical/U-VLM}.

\end{abstract}

\section{Introduction}

Automated radiology report generation for 3D medical imaging is key for reducing radiologist workload and improving diagnostic consistency. However, generating accurate reports requires multi-scale visual understanding: global context for anatomical regions, and fine-grained details for lesion detection. Existing 3D medical VLMs~\cite{hamamci2024ct2rep,hamamcibetter,bai2024m3d,hamamci2026generalist,wu2025towards,blankemeier2026merlin,shi2025medm} inject visual features only at the input layer of language models, losing multi-scale information during generation. Furthermore, no prior end-to-end VLM leverages dense per-voxel supervision from segmentation (Table~\ref{tab:param_compare})---despite SuPreM~\cite{li2024well} showing segmentation pretraining transfers more effectively than self-supervised approaches, and RadGPT~\cite{bassi2025radgpt} demonstrating segmentation-based methods outperform end-to-end VLMs in lesion detection (though it bypasses end-to-end learning via deterministic rules).

\begin{table}[t]
\centering
\caption{Comparison with related work on 3D radiology report generation. Existing methods inject visual features only at the language model input layer, and none leverage segmentation-pretrained U-Net for end-to-end report generation. VE: Vision Encoder. LD: Language Decoder. E2E: End-to-End. VE Init: CL (Contrastive Learning, CLIP~\cite{radford2021learning}-style), VQ Recon (vector-quantized reconstruction), Cls (classification), Seg (segmentation). Perc.: Perceiver. AP: Attention Pooling.}
\label{tab:param_compare}
\footnotesize
\setlength{\tabcolsep}{4pt}
\begin{tabular}{l|ccccc}
\toprule
\textbf{Method} & \textbf{VE} & \textbf{LD} & \textbf{VE Init} & \textbf{LD Init} & \textbf{E2E} \\
\midrule
CT2Rep~\cite{hamamci2024ct2rep} & Causal Trans. & Trans. Dec. & Scratch & Scratch & \ding{51} \\
RadFM~\cite{wu2025towards} & ViT+Perc. & MedLLaMA-13B & Scratch & Fine-tune & \ding{51} \\
M3D-LaMed~\cite{bai2024m3d} & ViT+Perc. & LLaMA-2-7B & CL & LoRA & \ding{51} \\
Merlin~\cite{blankemeier2026merlin} & 3D ResNet & RadLlama-7B & CL+Cls & LoRA & \ding{51} \\
CT-CHAT~\cite{hamamci2026generalist} & ViT+AP & LLaMA-3.1-70B & CL & LoRA & \ding{51} \\
MedM-VL~\cite{shi2025medm} & ViT+AP & Qwen2.5-3B & CL & Fine-tune & \ding{51} \\
BTB3D~\cite{hamamcibetter} & Causal Conv. & LLaMA-3.1-8B & VQ Recon & LoRA & \ding{51} \\
RadGPT~\cite{bassi2025radgpt} & U-Net Enc. & -- & Seg & -- & \ding{55} \\
\midrule
\textbf{U-VLM (Ours)} & U-Net Enc. & 0.1B & Seg+Cls & Scratch & \ding{51} \\
\bottomrule
\end{tabular}
\end{table}

U-Net~\cite{ronneberger2015u} dominates segmentation because its hierarchical encoder and skip connections preserve multi-scale information, with nnU-Net~\cite{isensee2021nnu,isensee2024nnu} providing a robust learning foundation. Existing 3D medical VLMs (Table~\ref{tab:param_compare}) lack this multi-scale hierarchy and inject visual features only at the language model input layer. ConvLLaVA~\cite{ge2024convllava} shows that hierarchical backbones produce over 4$\times$ fewer visual tokens than ViT at the same resolution while preserving rich visual information, motivating our hierarchical U-Net encoder; DeepStack~\cite{meng2024deepstack} shows that distributing visual tokens across multiple language model layers---rather than concentrating them at the input---enables models to handle 4$\times$ more visual tokens, motivating our multi-layer injection. We extend these insights to 3D medical imaging.

We propose U-VLM, a vision-language framework that enables hierarchical modeling in both training and architecture (Fig.~\ref{fig:framework}): progressive training from segmentation to classification to report generation first captures fine-grained spatial structures, then disease patterns, and finally supports report generation; multi-layer visual injection routes U-Net encoder features to corresponding language model layers, extending skip connections~\cite{ronneberger2015u} to vision-language modeling and preserving multi-scale information throughout generation.

To validate U-VLM, we train on CT-RATE~\cite{hamamci2026generalist} (25,692 chest CTs) and AbdomenAtlas 3.0~\cite{bassi2025radgpt} (9,262 abdominal CTs). U-VLM achieves F1 of 0.414 and BLEU-mean of 0.349 on CT-RATE, surpassing BTB3D (F1: 0.258, BLEU-mean: 0.305), and outperforms both end-to-end methods and segmentation-based detection on AbdomenAtlas (F1: 0.624 vs 0.518). U-VLM uses only a 0.1B decoder trained from scratch, while compared methods use 7B+ pre-trained models. Each training stage leverages different datasets without unified annotations. Our contributions (backed by Tables~\ref{tab:report_gen}--\ref{tab:ablation_report}) are: (1) a progressive pretraining pipeline from segmentation to classification to report generation; (2) multi-layer visual injection routing U-Net features to corresponding language model layers; (3) demonstrating that vision encoder pretraining outweighs the benefits of 7B+ pre-trained language models.

\section{U-VLM}

\begin{figure}[t]
    \centering
    \includegraphics[width=\textwidth]{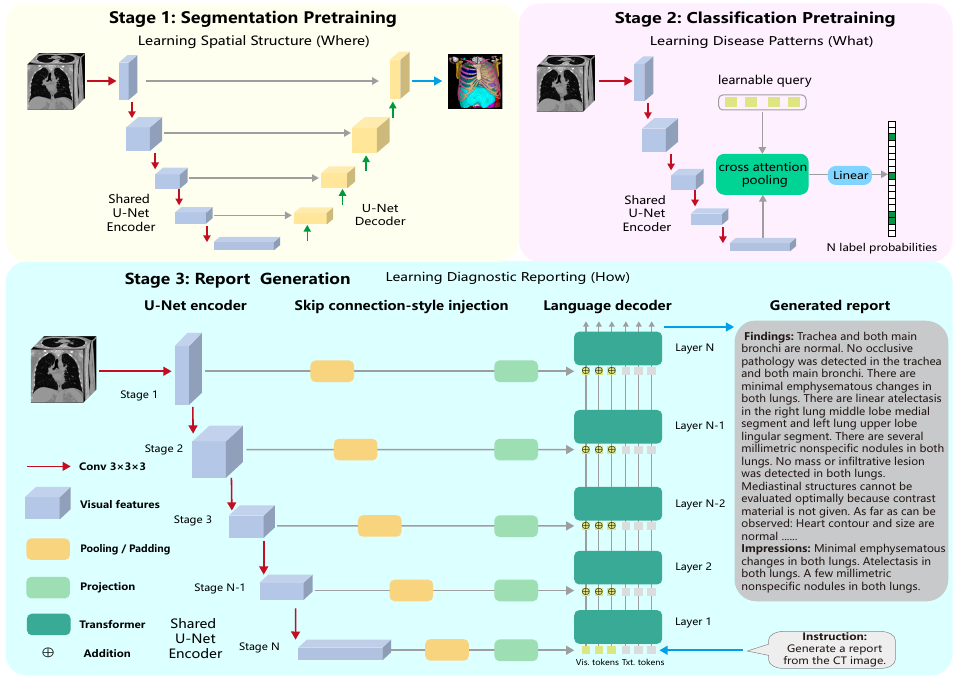}
    \caption{U-VLM framework. \textbf{Stage 1:} Segmentation pretraining for learning fine-grained spatial structures. \textbf{Stage 2:} Classification pretraining for disease pattern recognition. \textbf{Stage 3:} Report generation via multi-layer injection (deep encoder $\rightarrow$ early language layers, shallow encoder $\rightarrow$ later layers).}
    \label{fig:framework}
\end{figure}

U-VLM trains a shared U-Net encoder through three progressive stages, then connects it to a language decoder via multi-layer visual injection. Fig.~\ref{fig:framework} shows the framework overview.

\subsection{Progressive Training}
The shared U-Net encoder is sequentially optimized through three stages following curriculum learning~\cite{bengio2009curriculum}: Stage 1 learns spatial localization (``where'') from segmentation annotations, Stage 2 recognizes disease patterns (``what'') from classification labels, and Stage 3 generates reports (``how'') from image-report pairs. Each stage can leverage different datasets without unified annotations.

\textbf{Stage 1: Segmentation Pretraining.} The full U-Net learns fine-grained spatial structures through dense per-voxel supervision. We explore different segmentation granularities: (i) coarse anatomy only, (ii) coarse anatomy + lesions, and (iii) fine-grained anatomy + lesions, to analyze the impact on downstream report generation (Sec.~\ref{sec:ablation}). Given input volume $\mathbf{X} \in \mathbb{R}^{D \times H \times W}$ and ground-truth masks $\mathbf{Y}_{seg}$, the U-Net produces multi-scale encoder features $\{\mathbf{f}_i\}_{i=1}^N$ and decoder output $\hat{\mathbf{Y}}_{seg}$:
\begin{equation}
\mathcal{L}_{seg} = \mathcal{L}_{dice}(\mathbf{Y}_{seg}, \hat{\mathbf{Y}}_{seg}) + \mathcal{L}_{ce}(\mathbf{Y}_{seg}, \hat{\mathbf{Y}}_{seg})
\end{equation}

\textbf{Stage 2: Classification Pretraining.} The decoder is replaced with a classification head that uses learnable query vectors $\mathbf{Q} \in \mathbb{R}^{M \times D}$ to aggregate encoder features via cross-attention. Given multi-label disease annotations $\mathbf{y} \in \{0,1\}^C$:
\begin{equation}
\hat{\mathbf{y}} = \sigma(\text{Linear}(\text{Flatten}(\text{CrossAttn}(\mathbf{Q}, \mathbf{f}_N)))), \quad \mathcal{L}_{cls} = \text{BCE}(\mathbf{y}, \hat{\mathbf{y}})
\end{equation}

\textbf{Stage 3: Report Generation.} The pretrained encoder connects to a language decoder via multi-layer visual injection (Sec.~\ref{sec:injection}). The language modeling loss is:
\begin{equation}
\mathcal{L}_{gen} = -\sum_{t=1}^{T} \log P(w_t | w_{<t}, \{\mathbf{h}_j\}_{j=1}^L)
\end{equation}
where $\mathbf{h}_j$ denotes the hidden state at language layer $j$, enriched by injected visual features at vision token positions.

\subsection{Multi-Layer Visual Injection}
\label{sec:injection}
Standard VLMs typically inject visual features only at the input layer, causing multi-scale spatial details to vanish in deeper language layers. Following U-Net skip connections~\cite{ronneberger2015u} and DeepStack~\cite{meng2024deepstack}, we inject features from each encoder stage $S_i$ into specific language model layers $L_j$.

\textbf{Feature Alignment.} To enable injection across layers, features from different stages must share a consistent token sequence length $K$. We select a reference stage $S_r$ (where $1 \le r \le N$, typically $r=N-1$) to define $K$. For any stage $S_i$ with feature $\mathbf{f}_i$:
\begin{equation}
    \tilde{\mathbf{f}}_i = \text{Align}_r(\mathbf{f}_i) = \begin{cases}
    \text{Pool}(\mathbf{f}_i) & \text{if } i < r \\
    \mathbf{f}_i & \text{if } i = r \\
    \text{Pad}(\mathbf{f}_i) & \text{if } i > r
    \end{cases}
\end{equation}
Shallower stages ($i<r$) are downsampled via adaptive pooling, while deeper stages ($i>r$) are padded with zeros to match length $K$. A projection layer $\text{Proj}(\cdot)$ maps $\tilde{\mathbf{f}}_i$ to the language hidden dimension $D$.

\textbf{Skip Connection-Style Injection.} Analogous to U-Net's skip connections, we inject multi-scale features across language layers: deep encoder stages (global semantics) feed early language layers, while shallow stages (fine-grained details) feed later layers. Following DeepSeek-OCR 2~\cite{wei2026deepseek}, we adopt a hybrid attention mask where vision tokens attend bidirectionally while text tokens use causal attention. The hidden states at layer $j$ are updated as:
\begin{equation}
    \mathbf{h}_j^{(v)} = \text{LM}_j \left( \mathbf{h}_{j-1}^{(v)} + \text{Proj}_j \left( \text{Align}_r(\mathbf{f}_{N-j+1}) \right) \right), \quad \mathbf{h}_j^{(t)} = \text{LM}_j \left( \mathbf{h}_{j-1}^{(t)} \right)
\end{equation}
where visual features are injected only at vision token positions; text tokens pass through without injection. The mapping pairs encoder stage $N-j+1$ with language layer $j$ (e.g., $\mathbf{f}_N \rightarrow L_1$). The decoder has $N$ layers---one per encoder stage---yielding a one-to-one mapping ($\mathbf{f}_{N-j+1} \rightarrow L_j$ for $j=1,\dots,N$), so every language layer receives visual features from a distinct stage.

\section{Experiments}

\subsection{Datasets and Setup}
\label{sec:setup}
We evaluate U-VLM on two 3D CT datasets with no data leakage between training stages. (I) \textbf{CT-RATE}~\cite{hamamci2026generalist}: 25,692 chest CT volumes with reports and 18-class multi-label abnormality classification. For Stage 1, we use 2,628 cases from ReXGroundingCT~\cite{baharoon2026rexgroundingct} (a subset of CT-RATE training set) covering 14 lesion categories, augmented with pseudo-labels for anatomical structures from TotalSegmentator~\cite{wasserthal2023totalsegmentator} (33 classes), ATM22~\cite{zhang2023multi} (airway), and HiPaS~\cite{chu2025deep} (pulmonary vessels). Stages 2--3 use CT-RATE's native labels. (II) \textbf{AbdomenAtlas 3.0}~\cite{bassi2025radgpt}: 9,262 abdominal CT volumes with per-voxel lesion annotations, 38 fine-grained anatomy classes, structured reports, and 3-class lesion classification (liver, kidney, pancreas). All three stages use only AbdomenAtlas data (same as RadGPT), with no external datasets. Unlike RadGPT which reports tumor detection, we evaluate lesion detection since the publicly available AbdomenAtlas 3.0 does not include malignancy information.

\textbf{Implementation.} We implement U-VLM based on nnU-Net~\cite{isensee2021nnu,isensee2024nnu}. The U-Net encoder (ResEncoder) has 6 stages with features [32, 64, 128, 256, 320, 320]. Input volumes are resized to $256\times256\times192$; segmentation uses patches of $128\times128\times96$. The lightweight language decoder (0.1B parameters, 6 layers, 512 hidden dim, 8 heads) is trained from scratch, as task-specific lightweight models match adapted large models~\cite{du2025unirec}. For comparison, we also evaluate Qwen3-4B~\cite{yang2025qwen3} with both LoRA~\cite{hu2022lora} (rank 64, $\alpha$=128) and full parameter fine-tuning. All experiments use batch size 2 on a single NVIDIA A100 80GB GPU. Training uses SGD with learning rate 0.01 for segmentation (Stage 1), and AdamW with base learning rate 2e-5 for classification and report generation (Stages 2--3), where the encoder uses 0.1$\times$ multiplier when unfrozen or 0 when frozen. Metrics include BLEU-mean and macro-averaged F1, Precision, and Recall following CT-CHAT~\cite{hamamci2026generalist} and BTB3D~\cite{hamamcibetter}. Report-level clinical labels are extracted by task-specific evaluators: for CT-RATE, the official RadBERT text classifier (threshold 0.5) extracts the 18 abnormality labels from generated reports, applied identically to predictions and ground-truth reports; for AbdomenAtlas, a DeepSeek LLM fills a structured template (yes/no/uncertain for each organ) to extract liver, kidney and pancreas lesion labels, following the RadGPT evaluation protocol.

\subsection{Results}
\label{sec:ablation}

\begin{figure}[t]
    \centering
    \includegraphics[width=\textwidth]{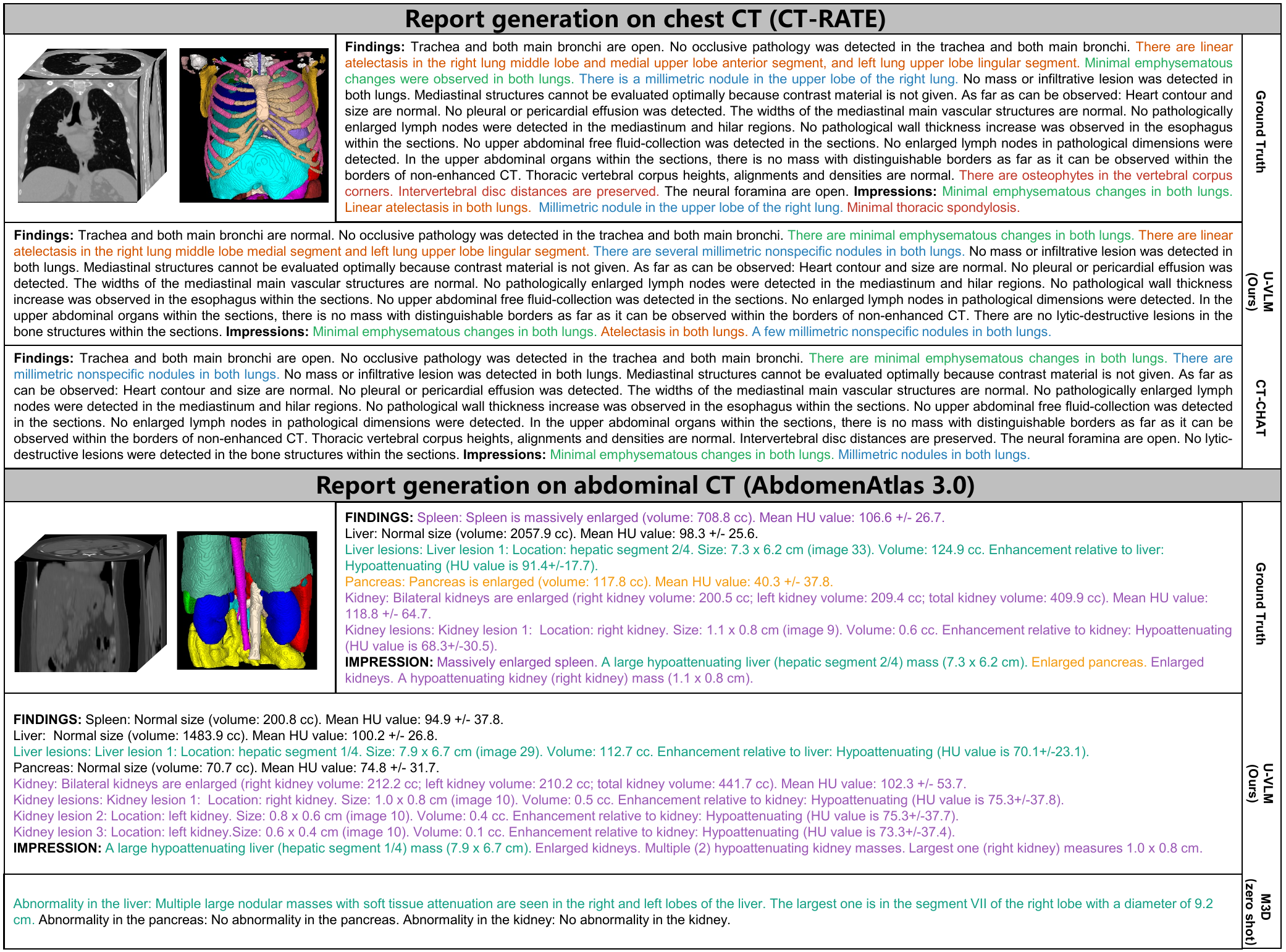}
    \caption{Qualitative results of segmentation and report generation on chest CT (CT-RATE) and abdominal CT (AbdomenAtlas 3.0). We visualize the input 3D CT volumes alongside segmentation predictions: Seg(F+L) for chest CT and Seg(C+L) for abdominal CT. For all reports, text is color-coded to highlight abnormalities, maintaining consistent colors for the same pathology, while normal descriptions are shown in black.}
    \label{fig:qualitative}
\end{figure}

\textbf{Main Results.} Table~\ref{tab:report_gen} compares U-VLM against existing methods on CT-RATE. U-VLM achieves F1 of 0.414, surpassing BTB3D-16 (0.258) by 60\% relative improvement, and BLEU-mean improves from 0.305 to 0.349---using only a 0.1B decoder trained from scratch, while compared methods use 7B+ pre-trained models.
Table~\ref{tab:lesion_iid} evaluates lesion detection on AbdomenAtlas 3.0. U-VLM with Seg(C+L) achieves best F1 across all organs (59.5\% pancreas, 64.8\% kidney, 62.9\% liver), outperforming both end-to-end methods (M3D, RadFM) and segmentation-based detection following RadGPT protocol~\cite{bassi2025radgpt} (nnU-Net). Fig.~\ref{fig:qualitative} shows qualitative results on both datasets.

\begin{table}[t]
\centering
\caption{Report generation on CT-RATE.}
\label{tab:report_gen}
\small
\setlength{\tabcolsep}{2.5pt}
\begin{tabular}{l|cccccccc}
\toprule
\textbf{Method} & \textbf{F1} & \textbf{Prec.} & \textbf{Rec.} & \textbf{B-1} & \textbf{B-2} & \textbf{B-3} & \textbf{B-4} & \textbf{B-mean} \\
\midrule
CT2Rep~\cite{hamamci2024ct2rep} & 0.160 & 0.435 & 0.128 & 0.372 & 0.292 & 0.243 & 0.213 & 0.280 \\
Merlin~\cite{blankemeier2026merlin} & 0.160 & 0.295 & 0.112 & 0.231 & 0.163 & 0.124 & 0.099 & 0.154 \\
CT-CHAT~\cite{hamamci2026generalist} & 0.184 & 0.450 & 0.158 & 0.373 & 0.284 & 0.231 & 0.198 & 0.272 \\
MedM-VL~\cite{shi2025medm} & 0.167 & 0.323 & 0.160 & 0.390 & 0.295 & 0.240 & 0.205 & 0.283 \\
BTB3D-8~\cite{hamamcibetter} & 0.187 & 0.260 & 0.150 & 0.411 & 0.307 & 0.245 & 0.215 & 0.295 \\
BTB3D-16~\cite{hamamcibetter} & \underline{0.258} & 0.260 & 0.260 & \underline{0.420} & \underline{0.318} & \underline{0.256} & \underline{0.225} & \underline{0.305} \\
\midrule
\textbf{U-VLM (Ours)} & \textbf{0.414} & 0.491 & 0.429 & \textbf{0.474} & \textbf{0.367} & \textbf{0.300} & \textbf{0.256} & \textbf{0.349} \\
\bottomrule
\end{tabular}
\end{table}

\begin{table}[t]
\centering
\caption{Lesion detection on AbdomenAtlas 3.0.}
\label{tab:lesion_iid}
\small
\setlength{\tabcolsep}{3pt}
\begin{tabular}{l|ccc|ccc|ccc}
\toprule
& \multicolumn{3}{c|}{\textbf{Pancreatic Lesion}} & \multicolumn{3}{c|}{\textbf{Kidney Lesion}} & \multicolumn{3}{c}{\textbf{Liver Lesion}} \\
\textbf{Model} & \footnotesize Prec. & \footnotesize Rec. & \footnotesize F1 & \footnotesize Prec. & \footnotesize Rec. & \footnotesize F1 & \footnotesize Prec. & \footnotesize Rec. & \footnotesize F1 \\
\midrule
\multicolumn{10}{l}{\textit{End-to-End Report Generation}} \\
M3D (zero-shot)~\cite{bai2024m3d} & 11.8 & 3.6 & 5.6 & 23.5 & 9.2 & 13.2 & 16.5 & 10.7 & 13.0 \\
RadFM (zero-shot)~\cite{wu2025towards} & 0.0 & 0.0 & 0.0 & 0.0 & 0.0 & 0.0 & 0.0 & 0.0 & 0.0 \\
\textbf{U-VLM Seg(C+L)} & 61.0 & 58.2 & \textbf{59.5} & 54.8 & 79.4 & \textbf{64.8} & 58.1 & 68.7 & \textbf{62.9} \\
U-VLM Seg(F+L) & 51.2 & 39.1 & 44.3 & 59.8 & 70.2 & \underline{64.6} & 45.8 & 75.6 & 57.1 \\
\midrule
\multicolumn{10}{l}{\textit{Segmentation-Based Detection (RadGPT-style)}} \\
nnU-Net Seg(C+L)~\cite{isensee2024nnu} & 69.1 & 50.9 & \underline{58.6} & 34.2 & 90.8 & 49.7 & 32.3 & 87.8 & 47.2 \\
nnU-Net Seg(F+L)~\cite{isensee2024nnu} & 86.7 & 35.5 & 50.3 & 49.1 & 74.8 & 59.3 & 49.2 & 48.1 & 48.6 \\
\bottomrule
\end{tabular}
\end{table}

\textbf{Ablation Studies.} Tables~\ref{tab:ablation_cls}--\ref{tab:ablation_report} analyze key design choices. \textit{Notations:} None: random initialization (no pretraining); Seg(C): coarse anatomy; Seg(C+L): coarse anatomy + lesions; Seg(F+L): fine-grained anatomy + lesions. B-M: BLEU-mean. SC: Skip Connections; Frz: Frozen Encoder; VT: Visual Tokens; Q4B-L/F: Qwen3-4B with LoRA/full fine-tuning. Classification metrics are extracted from generated reports by the evaluators described in Sec.~\ref{sec:setup}.
\textit{(1) Progressive training validates dense supervision.} The full Seg$\rightarrow$Cls$\rightarrow$Rep pipeline achieves F1=0.415 on CT-RATE, while skipping segmentation (Cls$\rightarrow$Rep: 0.292) or classification (Seg$\rightarrow$Rep: 0.238) degrades performance significantly. Table~\ref{tab:ablation_cls} shows segmentation pretraining improves classification by +23\% (CT-RATE) and +73\% (AbdomenAtlas), consistent with SuPreM~\cite{li2024well}. This confirms that dense per-voxel supervision from segmentation enables learning finer-grained spatial structures than training from scratch.
\textit{(2) Multi-layer injection preserves multi-scale information.} Skip connection-style injection improves B-M from 0.303 to 0.349 on CT-RATE and from 0.413 to 0.437 on AbdomenAtlas, while maintaining F1 (0.415$\rightarrow$0.414 and 0.620$\rightarrow$0.624), enhancing report fluency without sacrificing diagnostic accuracy. This supports our hypothesis that routing hierarchical features to corresponding language layers prevents information loss during generation.
\textit{(3) Segmentation granularity is task-dependent.} Optimal granularity varies by dataset: Seg(F+L) achieves best F1 on CT-RATE (0.415) where fine-grained localization aids 18-class classification, while Seg(C+L) yields better lesion detection on AbdomenAtlas (F1=0.624 vs 0.553) where fine-grained anatomy segmentation (e.g., liver segments, pancreas subdivisions) may distract from 3-class lesion classification.
\textit{(4) Encoder freezing preserves learned representations.} Frozen encoder (F1=0.415) outperforms fine-tuning (0.362), likely because freezing prevents catastrophic forgetting of discriminative features learned during classification pretraining.
\textit{(5) Visual token count shows diminishing returns.} Increasing tokens (384$\rightarrow$3072) yields no improvement, suggesting the bottleneck lies in other components rather than visual token quantity.
\textit{(6) Vision encoder pretraining outweighs decoder scale.} Our 0.1B decoder outperforms Qwen3-4B with both LoRA and full fine-tuning, consistent with UniRec~\cite{du2025unirec}. On CT-RATE, 0.1B achieves F1=0.415 vs Qwen3-4B's 0.167 (LoRA) and 0.156 (full); on AbdomenAtlas, the gap narrows (0.624 vs 0.596). Pre-trained LLMs may have difficulty adapting to small medical datasets with domain-specific output distributions.

\begin{table}[t!]
\centering
\caption{Classification ablation on CT-RATE and AbdomenAtlas 3.0.}
\label{tab:ablation_cls}
\small
\begin{tabular}{l|ccc|ccc}
\toprule
& \multicolumn{3}{c|}{\textbf{CT-RATE}} & \multicolumn{3}{c}{\textbf{AbdomenAtlas 3.0}} \\
\textbf{Training} & \textbf{F1} & \textbf{Precision} & \textbf{Recall} & \textbf{F1} & \textbf{Precision} & \textbf{Recall} \\
\midrule
None $\rightarrow$ Cls & 0.451 & 0.541 & 0.404 & 0.373 & 0.304 & 0.557 \\
Seg(C) $\rightarrow$ Cls & 0.546 & 0.587 & 0.535 & \underline{0.617} & 0.568 & 0.679 \\
Seg(C+L) $\rightarrow$ Cls & \underline{0.553} & 0.595 & 0.541 & \textbf{0.646} & 0.595 & 0.716 \\
Seg(F+L) $\rightarrow$ Cls & \textbf{0.555} & 0.598 & 0.539 & 0.578 & 0.567 & 0.589 \\
\bottomrule
\end{tabular}
\end{table}

\begin{table}[t!]
\centering
\caption{Report generation ablation on CT-RATE and AbdomenAtlas 3.0.}
\label{tab:ablation_report}
\footnotesize
\setlength{\tabcolsep}{2pt}
\begin{tabular}{l|cccc|cc|cc}
\toprule
& & & & & \multicolumn{2}{c|}{\textbf{CT-RATE}} & \multicolumn{2}{c}{\textbf{Abdomen}} \\
\cmidrule(lr){6-7} \cmidrule(lr){8-9}
\textbf{Training} & \textbf{SC} & \textbf{Frz} & \textbf{VT} & \textbf{Dec.} & \textbf{F1} & \textbf{B-M} & \textbf{F1} & \textbf{B-M} \\
\midrule
None $\rightarrow$ Rep & & \checkmark & 384 & 0.1B & 0.062 & 0.251 & 0.264 & 0.336 \\
Cls $\rightarrow$ Rep & & \checkmark & 384 & 0.1B & 0.292 & 0.294 & 0.384 & 0.391 \\
Seg(C+L) $\rightarrow$ Rep & & \checkmark & 384 & 0.1B & 0.232 & 0.297 & 0.348 & 0.341 \\
Seg(F+L) $\rightarrow$ Rep & & \checkmark & 384 & 0.1B & 0.238 & 0.278 & 0.398 & 0.392 \\
\cmidrule{1-9}
Seg(C) $\rightarrow$ Cls $\rightarrow$ Rep & & \checkmark & 384 & 0.1B & 0.399 & 0.298 & 0.592 & 0.423 \\
Seg(C+L) $\rightarrow$ Cls $\rightarrow$ Rep & & \checkmark & 384 & 0.1B & 0.409 & 0.279 & \underline{0.620} & 0.413 \\
Seg(F+L) $\rightarrow$ Cls $\rightarrow$ Rep & & \checkmark & 384 & 0.1B & \textbf{0.415} & 0.303 & 0.560 & 0.407 \\
\cmidrule{1-9}
Seg(C+L) $\rightarrow$ Cls $\rightarrow$ Rep & \checkmark & \checkmark & 384 & 0.1B & 0.407 & 0.298 & \textbf{0.624} & \textbf{0.437} \\
Seg(F+L) $\rightarrow$ Cls $\rightarrow$ Rep & \checkmark & \checkmark & 384 & 0.1B & \underline{0.414} & \textbf{0.349} & 0.553 & 0.417 \\
\cmidrule{1-9}
Seg(C+L) $\rightarrow$ Cls $\rightarrow$ Rep & & \checkmark & 3072 & 0.1B & 0.404 & 0.278 & 0.618 & 0.406 \\
Seg(F+L) $\rightarrow$ Cls $\rightarrow$ Rep & & \checkmark & 3072 & 0.1B & 0.413 & 0.286 & 0.563 & 0.398 \\
\cmidrule{1-9}
Seg(C+L) $\rightarrow$ Cls $\rightarrow$ Rep & & \checkmark & 384 & Q4B-L & 0.268 & 0.275 & 0.596 & 0.382 \\
Seg(C+L) $\rightarrow$ Cls $\rightarrow$ Rep & & \checkmark & 384 & Q4B-F & 0.275 & 0.288 & 0.588 & 0.391 \\
Seg(F+L) $\rightarrow$ Cls $\rightarrow$ Rep & & \checkmark & 384 & Q4B-L & 0.167 & 0.281 & 0.525 & 0.404 \\
Seg(F+L) $\rightarrow$ Cls $\rightarrow$ Rep & & \checkmark & 384 & Q4B-F & 0.156 & \underline{0.304} & 0.537 & 0.412 \\
\cmidrule{1-9}
Seg(C+L) $\rightarrow$ Cls $\rightarrow$ Rep & & & 384 & 0.1B & 0.374 & 0.289 & 0.592 & \underline{0.431} \\
Seg(F+L) $\rightarrow$ Cls $\rightarrow$ Rep & & & 384 & 0.1B & 0.362 & 0.285 & 0.536 & 0.429 \\
\bottomrule
\end{tabular}
\end{table}

\subsection{Limitations}

Our method has several limitations. First, precise numerical statements in generated reports (e.g., lesion dimensions, volumes, and attenuation values) are produced by the language decoder and are not explicitly grounded in image measurements, so they may be hallucinated; as shown in Fig.~\ref{fig:qualitative}, the generated report also misses or alters clinically important findings (e.g., splenic and pancreatic enlargement, and the number/laterality of renal lesions), which are not fully captured by F1 and BLEU. Second, the segmentation annotations for Stage 1 on CT-RATE are generated by automated models (e.g., the lesion pseudo-labels from ReXGroundingCT), so the quality of encoder pretraining is bounded by the accuracy of these annotations. We leave clinically grounded numerical estimation to future work.

\section{Conclusion}

U-VLM enables hierarchical vision-language modeling in both training and architecture for 3D radiology report generation. By leveraging different datasets at each stage without requiring unified annotations, U-VLM achieves state-of-the-art performance on CT-RATE (F1: 0.414 vs 0.258, BLEU-mean: 0.349 vs 0.305, Table~\ref{tab:report_gen}) and AbdomenAtlas 3.0 (F1: 0.624 vs 0.518 for segmentation-based detection, Table~\ref{tab:lesion_iid}) with only a 0.1B decoder trained from scratch. Our ablation studies (Tables~\ref{tab:ablation_cls}--\ref{tab:ablation_report}) show that progressive pretraining significantly improves F1, while multi-layer injection improves BLEU-mean---demonstrating that both dense supervision from progressive pretraining and skip connection-style visual injection yield complementary benefits for 3D medical vision-language tasks. This flexibility in leveraging diverse annotation types enables data aggregation across institutions, suggesting potential for scalable unified medical AI without costly unified labeling.

\begin{credits}
\subsubsection{\ackname} This work is supported by supported by National Key R\&D Program of China (2022ZD0212400), Science and Technology Commission of Shanghai Municipality (No.25511102602), Natural Science Foundation of China (62373243), and Shanghai Municipal Commission of Economy and Informatization (AI for Science Program: 2025-GZL-RGZN-BTBX-02023).

\subsubsection{\discintname}
The authors have no competing interests to declare that are relevant to the content of this article.
\end{credits}

\bibliographystyle{splncs04}
\bibliography{ref}

\end{document}